\documentclass[sigconf]{acmart}
\usepackage{multirow}
\usepackage{subfigure}
\usepackage{overpic}

\AtBeginDocument{%
  }

\setcopyright{acmlicensed}
\copyrightyear{2023}
\acmYear{2023}

\acmConference[MM '23] {Proceedings of the 31st ACM International Conference on Multimedia}{October 29--November 3, 2023}{Ottawa, ON, Canada.}
\acmBooktitle{Proceedings of the 31st ACM International Conference on Multimedia (MM '23), October 29--November 3, 2023, Ottawa, ON, Canada}
\acmPrice{15.00}
\acmISBN{979-8-4007-0108-5/23/10}
\acmDOI{10.1145/3581783.3611860}

\begin{document}

\title{StableVQA: A Deep No-Reference Quality Assessment Model for Video Stability}

\author{Tengchuan Kou}
\affiliation{%
 \institution{Shanghai Jiao Tong University}
 \city{Shanghai}
 \country{China}
 \orcid{0009-0001-1510-027X}
}
\email{2213889087@sjtu.edu.cn}

\author{Xiaohong Liu}
\authornotemark[1]
\affiliation{%
 \institution{Shanghai Jiao Tong University}
 \city{Shanghai}
 \country{China}}
\email{xiaohongliu@sjtu.edu.cn}

\author{Wei Sun}
\affiliation{%
 \institution{Shanghai Jiao Tong University}
 \city{Shanghai}
 \country{China}}
\email{sunguwei@sjtu.edu.cn}

\author{Jun Jia}
\affiliation{%
  \institution{Shanghai Jiao Tong University}
  \city{Shanghai}
  \country{China}
}
\email{jiajun0302@sjtu.edu.cn}

\author{Xiongkuo Min}
\affiliation{%
 \institution{Shanghai Jiao Tong University}
 \city{Shanghai}
 \country{China}}
\email{minxiongkuo@sjtu.edu.cn}

\author{Guangtao Zhai}
\affiliation{%
 \institution{Shanghai Jiao Tong University}
 \city{Shanghai}
 \country{China}}
\email{zhaiguangtao@sjtu.edu.cn} 

\author{Ning Liu}
\authornote{Corresponding authors.}
\affiliation{%
 \institution{Shanghai Jiao Tong University}
 \city{Shanghai}
 \country{China}}
\email{ningliu@sjtu.edu.cn}


\begin{abstract}
  Video shakiness is an unpleasant distortion of User Generated Content (UGC) videos, which is usually caused by the unstable hold of cameras. In recent years, many video stabilization algorithms have been proposed, yet no specific and accurate metric enables comprehensively evaluating the stability of videos. Indeed, most existing quality assessment models evaluate video quality as a whole without specifically taking the subjective experience of video stability into consideration. Therefore, these models cannot measure the video stability explicitly and precisely when severe shakes are present. In addition, there is no large-scale video database in public that includes various degrees of shaky videos with the corresponding subjective scores available, which hinders the development of Video Quality Assessment for Stability (VQA-S). To this end, we build a new database named \textbf{\textit{StableDB}} that contains $1,952$ diversely-shaky UGC videos, where each video has a Mean Opinion Score (MOS) on the degree of video stability rated by 34 subjects. Moreover, we elaborately design a novel VQA-S model named \textbf{\textit{StableVQA}}, which consists of three feature extractors to acquire the optical flow, semantic, and blur features respectively, and a regression layer to predict the final stability score. Extensive experiments demonstrate that the StableVQA achieves a higher correlation with subjective opinions than the existing VQA-S models and generic VQA models. The database and codes are available at \href{https://github.com/QMME/StableVQA}{https://github.com/QMME/StableVQA}.

\end{abstract}



\maketitle

\section{Introduction}

\begin{figure}
    \centering
    \begin{overpic}[width=\linewidth]{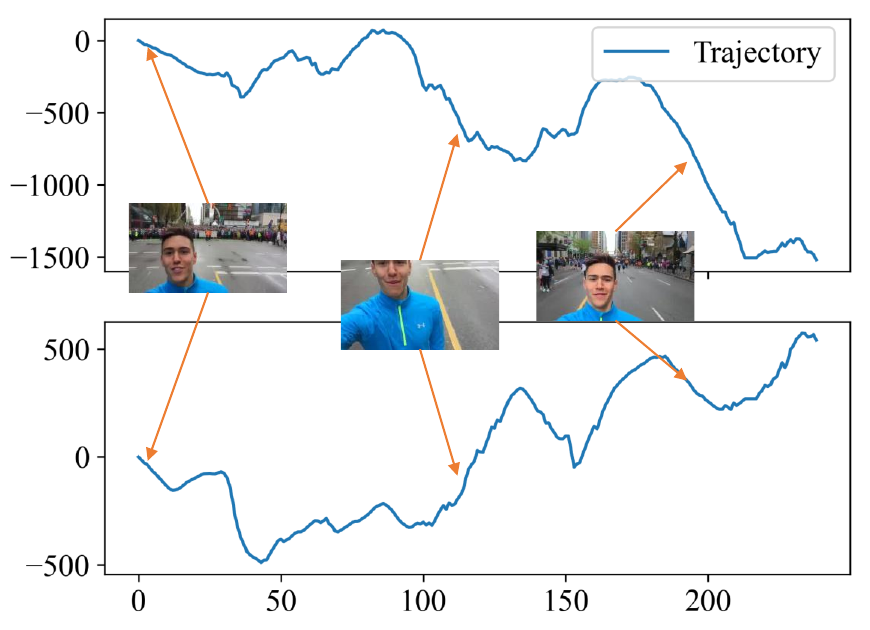}
    \put(45, -3){Frame Number}
    \put(-2, 20){\rotatebox{90}{\small{dy}}}
    \put(-2, 50){\rotatebox{90}{\small{dx}}}
    \end{overpic}
    \vspace{2mm}
    \caption{Trajectory of a sample video on x and y axes, whose stability score~\cite{liu2013bundled} is 84.2 while subjective MOS is 44.6 (range 0 to 100 from unstable to stable).} 
    \label{fig:sample}
\end{figure}

With the development of streaming media such as YouTube, TikTok, etc., people receive a huge amount of User Generated Content (UGC) videos daily. 
UGC videos generally suffer from significant distortions such as blurring, low light, and shakiness, which decreases the Quality of Experience (QoE) of viewers. In these distortion categories, shakiness is the most unpleasant one that may dizzy the viewers, which is caused by the unstable hold of cameras. In the temporal dimension, the unstable movement of cameras forms a shaky pixel trajectory as shown in Fig.~\ref{fig:sample}. 
However, although plenty of video stabilization algorithms are proposed to remove shakiness in recent years~\cite{liu2013bundled, james2023globalflownet, choi2020deep, zhao2020pwstablenet, yu2020learning}, to the best of our knowledge, there is no specific and accurate metric to evaluate video stability. 




Evaluating video stability is one of the essential factors in Video Quality Assessment (VQA), yet has not received extensive attention. 
According to the provided amount of pristine video information, VQA models can be categorized as No-Reference (NR), Reduced-Reference (RR), and Full-Reference (FR) models~\cite{sun2022deep}. 
Since the acquisition of stable reference videos relies on the support of stabilizers and synchronous shooting, making the FR Video Quality Assessment for Stability (VQA-S) is still impractical. Therefore, most existing VQA-S models focus on NR quality assessment.

Among all VQA-S models, Stability Score~\cite{liu2013bundled} is widely used to evaluate the performance of video stabilization algorithms. 
Stability Score follows the principle that the more energy contained in the low-frequency parts, the more stable a video is. However, experimental results demonstrate that the Stability Score may not reflect the subjective experience in video stability. Specifically, it estimates the camera movement by using feature point matching to compute the homography between adjacent frames. When the feature points are located in moving objects whose trajectory is not consistent with the camera, even though the camera is stable, it still regards the video as being shaky. In addition, when the video suffers from severe shakiness, the caused motion blur would impede the point matching. Figure~\ref{fig:sample} gives an example of when the Stability Score fails to predict correctly. 


In recent years, a variety of VQA databases have been proposed. Benefiting from them, the VQA models based on Deep Neural Networks (DNN) have emerged. However, the tailored quality assessment database for video stability is still lacking, which hinders the development of effective VQA-S models.
To this end, we build \textbf{\textit{StableDB}}, a database with 1,952 diversely-shaky videos in various in-the-wild scenes, and conduct a subjective study on 34 subjects to obtain the corresponding MOS. To the best of our knowledge, this is the largest video database which contains different shake degrees to support the accurate assessment of video stability. 
We anticipate the StableDB would benefit the training and testing of subsequent models.




On the base of StableDB, we propose \textbf{\textit{StableVQA}}, a novel DNN-based VQA-S model that integrates three tailored features to better assess the shake degrees in videos. 
The proposed StableVQA consists of feature extraction, feature fusion, and quality regression modules. 
Concretely, we extract features from optical flow, semantic domain, and blur domain respectively. The optical flow explicitly describes pixel movement between frames. A 3D-CNN is used to implicitly analyze the camera movement within optical flows, which is significant for the assessment of video stability. A Swin Transformer~\cite{wu2022fast} backbone is used for extraction of semantic features to help with eliminating effects from moving objects whose trajectories are inconsistent with the camera movement. Besides, a blur encoder is designed to detect the blur effect caused by the high-speed movement of the camera within frames. 
We train and test StableVQA as well as other VQA models on StableDB and other public databases. Experimental results show the StableVQA outperforms the existing VQA-S models and the state-of-the-art VQA model, validating its effectiveness in measuring video stability. 

We summarize our contributions as follows:

\begin{enumerate}
    \item We build the first large-scale subjective video database containing videos of various shake degrees, named StableDB. 
    The database includes 1,952 video sequences and corresponding MOSs on video stability gained from 34 subjects. 
    
    \item We propose the first DNN-based model to predict 
    degrees of video stability, named StableVQA. 
    We creatively extract three tailored features in optical flow, semantic domain, and blur domain to benefit the evaluation of video stability. With the following feature fusion and quality regression modules, the model is able to predict video stability with high consistency to subjective opinions.
    
    
    \item The proposed StableVQA outperforms existing VQA-S models and the state-of-the-art VQA model on StableDB and public databases, indicating the effectiveness of StableVQA. Qualitative experiments show StableVQA can benefit in measuring the performance of video stabilization algorithms, giving it practical application prospects.

\end{enumerate}

\begin{table*}[]
\centering
\caption{Summary of StableDB and the utilized databases. }
\renewcommand{\arraystretch}{1.2}
\begin{tabular}{c|ccccccllll}
\toprule
    & Name    & Total Videos   & Sampled Videos & Resolution & Time Duration & Video Format\\
    \cline{1-7}
\multirow{4}{*}{UGC databases}     
    & KoNViD-1k~\cite{hosu2017konstanz}  & 1200  & 854  & 540p  & 8s &   MP4\\
    & V3C1~\cite{rossetto2019v3c}  & 7475  & 700  & 176p-4K   & 3-60min &   MP4    \\
    & LIVE-Qualcomm~\cite{ghadiyaram2017capture}  & 208        & 54    & 1080p   & 15s  & YUV \\
    & YouTube UGC~\cite{wang2019youtube}  & 1500   & 39        & 360p-4K     & 20s   & YUV, MP4  \\ 
    \cline{1-7}
\multirow{3}{*}{Unstable databases} 
    & NUS~\cite{liu2013bundled}    & 174    & 105   & 360p, 720p    & 10-60s  &  AVI  \\
    & DeepStab~\cite{wang2018deep}   & 122     & 108     & 720p    &  6-75s   &   AVI    \\
    & Selfie~\cite{yu2018selfie}    & 33     & 22   & 480p     & 4-13s    &  MP4 \\ 
    \cline{1-7}
    Proposed     & StableDB  & 1952  & -   & 720p   & 8s   & MP4 \\ \bottomrule

\end{tabular}
\label{tab:databases}
\end{table*}

\section{Related Works}

\subsection{Unstable Video databases}


Several unstable video databases have been proposed for evaluating video stabilization algorithms, including NUS~\cite{liu2013bundled}, DeepStab~\cite{wang2018deep}, Selfie~\cite{yu2018selfie}, etc. However, the aforementioned unstable video databases are neither in abundance in amount nor with corresponding subjective opinion scores on stability. A large-scale subjective unstable video database is urged for the convenience of deep learning training.


There are also several UGC video databases for genetic VQA tasks, including KoNViD-1k~\cite{hosu2017konstanz}, V3C1~\cite{rossetto2019v3c}, LIVE-Qualcomm~\cite{ghadiyaram2017capture}, YouTube UGC~\cite{wang2019youtube}, VDPVE~\cite{gao2023vdpve}, etc.
Though the above large-scale UGC databases have been proposed, they may not be suitable for measuring stability. For one reason, partial UGC videos suffer from severe distortions from other dimensions such as blur, low light, high contrast, etc. It could affect the QoE of viewers in that they cannot focus on the stability of videos. On the other hand, common UGC videos from online platforms have transitions, which may lead to ambiguous definitions of stability.   

Moreover, there exists an absence of subjective study for measuring the degree of video shakiness specifically. Relative comprehensive studies are mostly designed for the comparison of video stabilization algorithms. In~\cite{wang2016pixel}, subjects were asked to rate a score from 0 to 100 based on video stability. However, the experiment was only conducted on 10 sets of videos with one shaky video and three stabilized videos using different stabilization models in one set. In~\cite{liu2013bundled, zhang2015simultaneous, koh2015video, xu2022dut}, the original shaky video was displayed along with the corresponding stabilized video. Subjects were required to choose the better results. Hence we propose a large-scale unstable video database with corresponding Mean Opinion Scores (MOS) to eliminate the gap in scale and subjective study. 

\subsection{No-reference VQA}

Since it is impractical to obtain stable reference video pairing with a target unstable video, a no-reference quality assessment model is necessary for measuring video stability. A few VQA-S metrics have been proposed. In~\cite{battiato2007sift}, Interframe Transformation Fidelity (ITF) was used for assessing stability. It was calculated by averaging PSNR between adjacent frames. In~\cite{liu2013bundled}, Liu \textit{et al}. proposed to measure overall video stability by extracting the low-frequency component from camera movement. It was believed that the more energy contained in the low-frequency part, the more stable a video was. Both models fail to predict accurately when facing severe shaky motions. In recent years, Zhang \textit{et al}.~\cite{zhang2018intrinsic} proposed to mathematically analyze the intrinsic smoothness of the motion path. However, all the above models only consider one aspect of video stability while ignoring the subjective experience~\cite{guilluy2021video}. 

Most VQA models consider videos' overall quality but do not specifically focus on stability. ~\cite{mittal2012making, mittal2012no, ye2012unsupervised, sun2019mc360iqa} utilize NR Image Quality Assessment (IQA) models on frames of videos and pool the results as the video quality score, while ~\cite{mittal2015completely, korhonen2019two, tu2021ugc} utilize tailored handcrafted features for assessing video quality.

With the thriving of deep learning, numerous neural network-based VQA models have been proposed. VSFA~\cite{li2019quality} extracted semantic features from a pre-trained CNN model while using a gated recurrent unit network to model the temporal-memory effects. Li \textit{et al}.~\cite{li2022blindly} proposed to transfer knowledge from IQA databases by extracting spatial features using a pre-trained model. 
Sun \textit{et al}.~\cite{sun2022deep} trained a spatial feature extractor with the help of motion features extracted from pre-trained CNN. 
Zhang \textit{et al.}~\cite{zhang2023md} proposed MD-VQA, to measure the visual quality of UGC live videos from semantic, distortion, and motion aspects respectively.
Wu \textit{et al}.~\cite{wu2022fast} proposed a new sample strategy called ``fragment'' in FAST-VQA. It considered local quality and global quality with mini-patches sampled in uniform grids. 
Dong \textit{et al}.~\cite{dong2023light} proposed Light-VQA for the assessment of low-light video enhancement algorithms. Recently, there are several models aiming to evaluate a specific type of content, such as Artificial Intelligence Generated Content (AIGC) image~\cite{zhang2023perceptual, li2023agiqa}, 4K content~\cite{lu2022deep} and digital human~\cite{zhang2023advancing}.
However, the aforementioned models are not able to accurately assess video stability specifically, as they only consider stability as a factor in the overall video quality.

\section{Database Preparation and Subjective Study}
\label{database}

The existing unstable video databases, such as NUS~\cite{liu2013bundled}, DeepStab~\cite{wang2018deep}, and Selfie~\cite{yu2018selfie}, contain relatively few numbers of videos as shown in Table~\ref{tab:databases}, which cannot satisfy the training of DNN-based model. In addition, video shakiness is a common phenomenon in UGC videos and small-scale databases cannot represent the complicated scenes of UGC videos. Consequently, we propose a large-scale video database, named StableDB, including 1,952 UGC videos with various shaky degrees. Furthermore, we conduct a subjective study on 34 subjects to obtain the mean opinion score (MOS) of each video. In this section, we will describe the construction of StableDB and the subjective experiment conducted on it. 

\subsection{Data Acquisition}

\textbf{Data Sources:} Sources of StableDB include: (1) videos from existing video databases, (2) videos shot by ourselves. Videos in the existing unstable video databases are naturally suitable for StableDB. To enlarge the scale of StableDB, we turn to utilize the existing UGC video databases for VQA task, including KoNViD-1k~\cite{hosu2017konstanz}, V3C1~\cite{rossetto2019v3c}, LIVE-Qualcomm~\cite{ghadiyaram2017capture}, and YouTube UGC~\cite{wang2019youtube}. However, in addition to video shakiness, various other kinds of distortions are discovered in part of these videos, such as blur, low light, and high contrast. To reduce the influence of these irrelevant distortions, we manually select videos with broadly similar levels of other distortions but various shaky effects. In addition to these existing videos, we capture 70 video sequences using Apple iPhone 11. 

\textbf{Data Preparation:} We set each video sequence with duration of $8s$ and resize to the resolution of 720p. To avoid an ambiguous degree of motion shakiness, the video sequence should only contain a single shot without changing shooting scenes. Finally, we obtain 1,952 videos with detailed numbers listed in Table~\ref{tab:databases}.

\subsection{Subjective Study Design}

To obtain the actual situation of videos' stability in StableDB, we conduct a subjective study where subjects are required to rate on the degree of shakiness in videos based on subjective experience. The subjective study consists of a pilot study and the formal study. In the pilot study, we invite 8 volunteers to score 100 videos to have a general understanding of the data situation. In the formal study, we invite 34 volunteers to score the degree of shakiness of 1,952 videos, and finally obtain 66,368 ratings. Subjects are first shown an interface with the introduction of overall study settings. Following is the training session, where subjects get familiar with study settings and operations by scoring 5 videos with a wide range of stability. Afterward, subjects enter the testing session and rate each video of StableDB. All subjects are crowd-sourced.
We provide the detailed overall study workflow in Appendix~\ref{subjective}. 

\begin{figure}[]
  \centering
  \includegraphics[width=\linewidth]{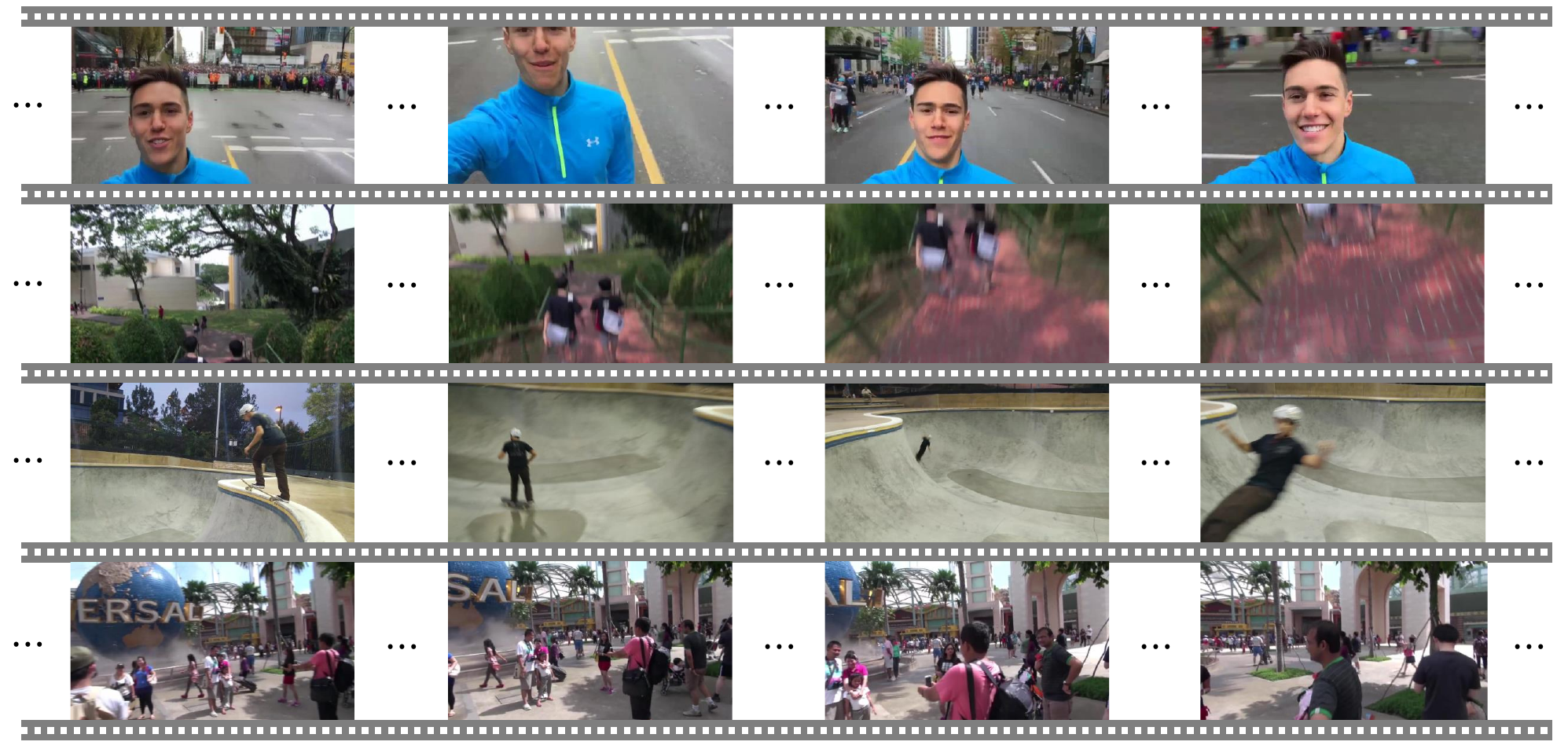}
  \caption{Sample video frames from StableDB.}
\end{figure}





\subsection{Quality Control} 


\begin{figure}[]
\centering
\subfigcapskip=7pt
\subfigbottomskip=-1pt
\subfigure[]{
\begin{minipage}[t]{0.5\linewidth}
\centering
\begin{overpic}[width=\linewidth]{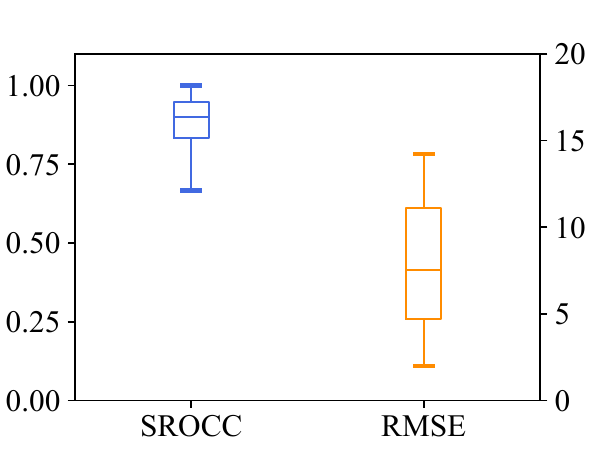}
\end{overpic}
\end{minipage}%
\label{fig:control}
}%
\subfigure[]{
\begin{minipage}[t]{0.5\linewidth}
\centering
\begin{overpic}[width=\linewidth]{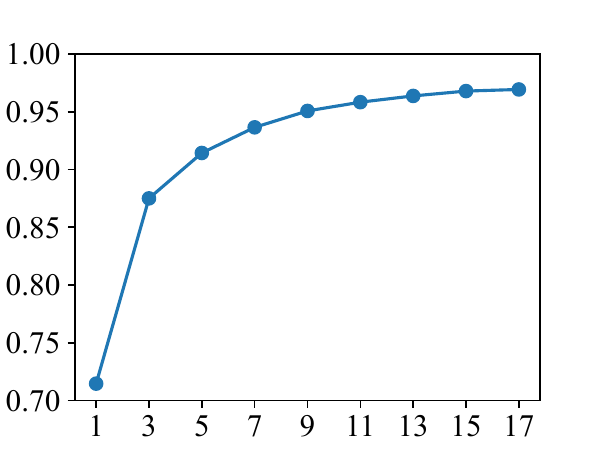}
\put(50, -5){\small{n}}
\end{overpic}
\end{minipage}
\label{fig:split_srocc}
}\\%
\caption{(a): Distribution of subjects' SROCC with golden videos and RMSE between repeated videos. SROCC refers to the left y-axis, and RMSE refers to the right. (b): SROCC between split groups.}
\end{figure}

When conducting the subjective study, quality control is necessary to obtain reliable results. Following~\cite{ying2022telepresence}, we conduct quality control from two aspects:

\textbf{Golden Videos:} Using ``golden videos'' for quality control is a common strategy in subjective studies. The golden videos are selected to have the most consistent scores in the pilot study. We compute the Spearman Rank Order Correlation Coefficient (SROCC) between the subject's scores on golden videos in testing and the MOSs in the pilot study. SROCC close to 1 means high reliability of the subject's scores. 

\textbf{Repeated Videos:} 5 randomly selected videos are repeated in both sessions during testing. We calculate the Root Mean Square Error (RMSE) between the subject's first and second ratings. A lower RMSE means a higher intra-subject consistency. 

The results of golden videos and repeated videos are shown in Figure~\ref{fig:control}. The average SROCC is 0.8806 with a standard deviation of 0.1002, while the average RMSE is 7.8940 with a standard deviation of 3.8523. Together they guarantee the reliability and effectiveness of the subjective study. 

Besides, to further prove the reliability of the subjective study, we conduct an experiment where we randomly select $2n$ subjects and equally split them  into 2 groups. After having the MOSs in each group, we calculate the SROCC between the two groups to evaluate the consistency. A higher SROCC means higher consistency in the two groups and reflects the reliability of the subjective experiment. We have in total 34 subjects, so the number of subjects in each group $n$ increases from 1 to 17. For each $n$, the procedure is repeated for 100 times, and we take the average SROCC. The results are shown in Figure~\ref{fig:split_srocc}. It can be observed that the gain in SROCC becomes minimal as the number 
 increases, indicating reliable ground truth can be obtained at the current subjective study scale.

\section{Proposed Model}

\begin{figure*}
    \centering
    \includegraphics[width=\textwidth]{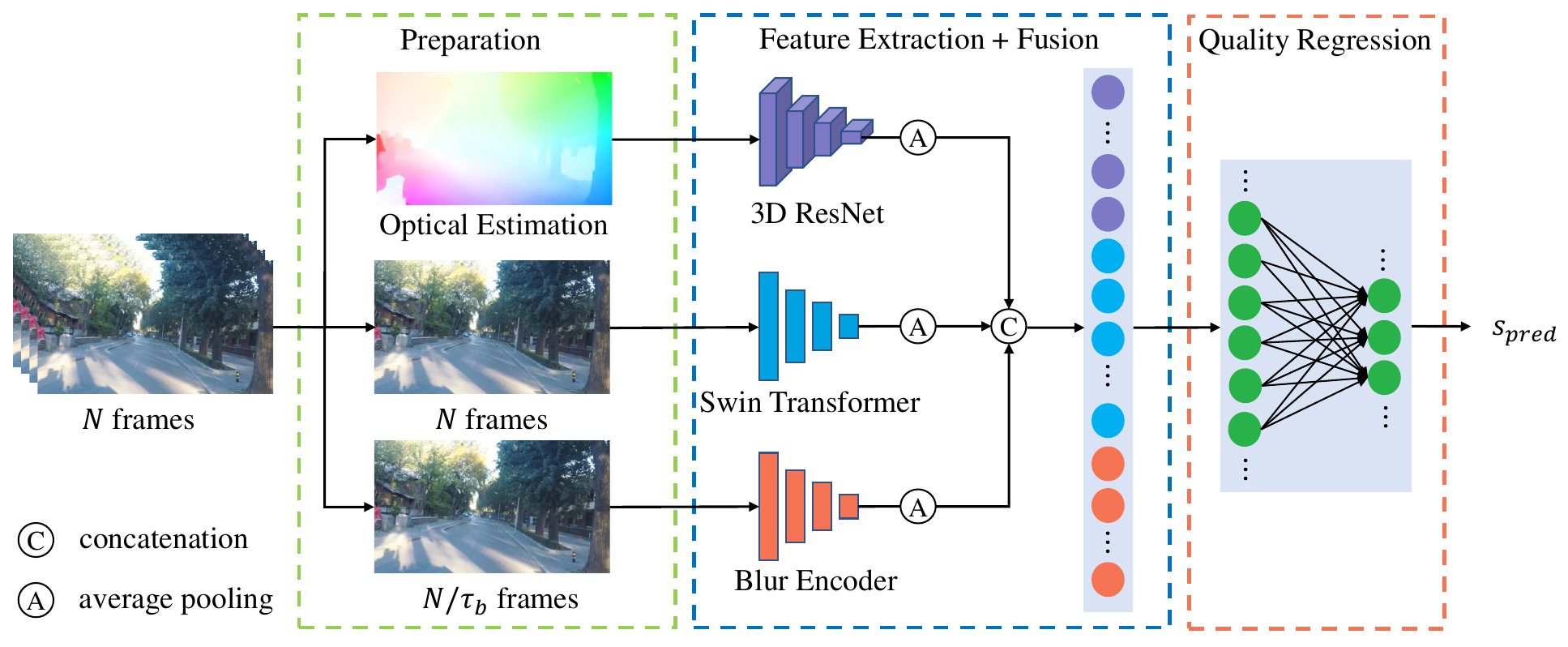}
    \caption{The overview of StableVQA. Features are extracted from optical flow, semantic domain, and blur domain. After fusion, A two-layer MLP is utilized to regress the concatenated feature to the predicted stability score.}
    \label{fig:model}
\end{figure*}


Based on StableDB, we propose a novel no-reference DNN-based model to evaluate video stability, named StableVQA. Figure~\ref{fig:model} gives an overview of StableVQA. The framework consists of three modules: feature extraction, feature fusion, and quality regression. We first randomly sample $N$ frames with time interval $\tau$ as a video clip. The optical flows between adjacent frames in the clip are estimated and are taken as input to a 3D Convolutional Neural Network (CNN) to implicitly analyze the camera movement as the flow feature. Besides, we use a Swin Transformer~\cite{liu2021swin} for the extraction of semantic features. Last but not least, we analyze the motion blur effect within frames as the blur feature. Features from these three dimensions are fused and regressed to give the final prediction stability score. The following introduces the detailed design of StableVQA.

\subsection{Preparation}


Jitter effects in video originate from the rough trajectory of the shooting equipment, which can be reflected by pixel motion in frames at the temporal dimension. Optical flow estimates the instantaneous velocity of moving objects at the pixel level. It finds the corresponding relationship between the previous frame and the current frame by using the changes of pixels in the time domain and the correlation between adjacent frames, so as to calculate the motion information of objects between adjacent frames. The analysis in optical flow will directly benefit the assessment of video stability. Given a video clip with $N$ frames, denoted $i = \{i_n\}^{N}_{n=1}$. we estimate the optical flows between adjacent frames in $i$ as the optical field, denoted as $o = \{o_n\}^{N-1}_{n=1}$, where 

\begin{equation}
    o_n = OPTICAL(i_n, i_{n+1}).
\end{equation}

\subsection{Feature Extraction}

Here we describe the details of the flow feature, semantic feature, and blur feature extraction respectively. We denote $C$, $T$, $H$ , and $W$ as the channel number, temporal length, spatial height, and spatial width of the feature map.

\subsubsection{Flow Feature}

 The optical field explicitly demonstrates frames-wise motion at the pixel level. Hence a 3D CNN is trained for the extraction of the implicit feature. Compared with 2D CNN, 3D CNN has the advantage of analyzing temporal information in videos. Given the optical field $o \in \mathbb{R}^{2 \times (N-1) \times H \times W }$, the extracted flow feature is noted as $f_o(o) \in \mathbb{R}^{C_o \times T_o \times H_o \times W_o}$.
 The trainable CNN model is noted as $f_o(\cdot)$. We use $f_o$ in replacement of $f_o(o)$ for simplicity.    

\subsubsection{Semantic Feature}

As the flow feature gives a video's characteristics in the temporal domain, we also analyze the video's semantic feature in the image field. Since moving objects in video scenes, like humans, vehicles, etc, often have trajectories inconsistent with the camera movement, semantic analysis will benefit in eliminating the impact of moving objects. 
Swin transformer~\cite{liu2021swin} has proved its efficiency in multiple vision tasks. It consists of four hierarchical self-attention layers with a shifted windowing scheme. Self-attention computations are done in non-overlapping local windows while the shifted windowing scheme guarantees the cross-window connection. Here we utilize Swin-T as the backbone of the semantic extraction model, denoted as $SWIN(\cdot)$. Given the video clip $i = \{i_n\}^{N}_{n=1}$, the semantic feature is given by:

\begin{equation}
    f_s(i) = cat(\{SWIN(i_{n})\}^{N}_{n=1}),
\end{equation}
where $cat$ denotes concatenation, as we concatenate frame-wise features in channel dimension. Given the image field $i \in \mathbb{R} ^ {N \times 3 \times H \times W}$ of a video, we have semantic feature $f_s \in \mathbb{R} ^ {(N \times C_s) \times H_s \times W_s}$.

\subsubsection{Blur Feature}

In the process of capturing in-the-wild videos, exposure time is a necessary factor to take into consideration. Exposure time stands for the certain time required to sensitize a photographic plate. If the shooting equipment suffers from severe shakiness, objects in scenes are still with great motion in one exposure time, resulting in motion blur in videos. The degree of motion blur reflects the video's stability from another perspective. Therefore, we utilize a pre-trained encoder in a typical image deblurring network, as it is designed for analyzing the blurring effect within images. Given the video clip $i = \{i_n\}^{N}_{n=1}$, we sample $N_b$ frames with time interval $\tau_b$ for blur detection, denoted as $i^b = \{i_{n \cdot \tau_b}\}^{N_b}_{n=1}$, where $N_b = N/\tau_b$. The blur feature is given by: 

\begin{equation}
    f_b(i^b) = cat(\{BLUR(i^b_n)\}^{N_b}_{n=1}),
\end{equation}
where $BLUR(\cdot)$ denotes the encoder of the utilized deblurring network. Similar to semantic features, we concatenate frame-wise features in channel dimension. Given $i^b \in \mathbb{R} ^ {N_b \times 3 \times H \times W}$, we have the blur feature $f_b \in \mathbb{R} ^ {(N_b \times C_b) \times H_b \times W_b}$.

\begin{table*}[]
\centering
\renewcommand{\arraystretch}{1.2}
\caption{Performance of the SOTA models and StableVQA on StableDB. The best-performing model is highlighted in each column. [Keys: of: flow feature; bf: blur feature]}
\begin{tabular}{c|c|cccc|cccc}
\toprule
\multirow{2}{*}{Type} & \multirow{2}{*}{models} & \multicolumn{4}{c|}{Validation}    & \multicolumn{4}{c}{Testing}        \\ \cline{3-10} 
                      &                          & SROCC $\uparrow$  & PLCC $\uparrow$ & KRCC $\uparrow$ & RMSE $\downarrow$ & SROCC $\uparrow$ & PLCC $\uparrow$& KRCC $\uparrow$ & RMSE $\downarrow$ \\ \hline
\multirow{2}{*}{VQA-S} & ITF~\cite{battiato2007sift}                     & 0.6148 & 0.5937 & 0.4366 & 15.3144 & 0.6138 & 0.5985 & 0.4353 & 15.2121 \\
                      & Stability Score~\cite{liu2013bundled}                      & 0.2365 & 0.2783 & 0.1587 & 20.4064 & 0.2217 & 0.2818 & 0.1489 & 20.363  \\ \hline
\multirow{6}{*}{VQA}  & VSFA~\cite{li2019quality}                    & 0.6516 & 0.6712 & 0.4726 & 13.2789 & 0.6166 & 0.6565 & 0.4465 & 13.6174 \\
                      & SimpleVQA~\cite{sun2022deep}                & 0.6368 & 0.6667 & 0.4599 & 12.7832 & 0.6285 & 0.6418 & 0.4753 & 12.8383 \\
                      & BVQA~\cite{li2022blindly}                & 0.8734 & 0.8774 & 0.7014 & 8.0137  & 0.8715 & 0.8767 & 0.6934 & 8.1371  \\
                      & FAST-VQA~\cite{wu2022fast}                & 0.8886 & 0.8908 & 0.7153 & 8.0055  & 0.8816 & 0.8873 & 0.7079 & 8.1594  \\
                      & FAST-VQA + of                & 0.8883 & 0.8867 & 0.7050 & 8.3924  & 0.8857 & 0.8833 & 0.7091 & 8.1254  \\
                      & FAST-VQA + bf                & 0.8892 & 0.882 & 0.7169 & 8.4943  & 0.8787 & 0.887 & 0.6988 & 7.9932  \\ \hline
\textbf{VQA-S}                       
              & \textbf{StableVQA}                     & \textbf{0.9102}        &  \textbf{0.9161}     & \textbf{0.7431}      & \textbf{7.0188}      & \textbf{0.9118}       & \textbf{0.9187}     & \textbf{0.7441}     & \textbf{6.9364}  \\ \bottomrule
\end{tabular}
\label{tab:performance}
\end{table*}

\subsection{Feature Fusion}

After feature extraction, we propose a feature fusion module to obtain an overall feature of the video. Firstly, adaptive average pooling is deployed for all three branches to unify features in the channel dimension. Then the overall feature is obtained by a concatenation:

\begin{equation}
    f = cat(avg\_pool(f_o, f_s, f_b)),
\end{equation}
where $f \in \mathbb{R}^{(C_o + N \times C_s + N_b \times C_b) \times 1}$. It is worth noting that a variety of fusion strategies can be leveraged for feature fusion, \textit{e.g.,} attention-based. However, they are beyond the scope of this paper. 

\subsection{Quality Regression}

Given the fused feature of the source video, a regression module is needed to map the feature representation to the quality score. Here we utilize a two-layer multi-layer perception (MLP) module for quality scores regression. The MLP consists of two fully connected layers and there are 128 and 1 neuron in each layer respectively. The predicted stability score is given by:

\begin{equation}
    s_{pred} = FC(f),
\end{equation}
where $FC(\cdot)$ denotes the function of the two fully connected layers.

\section{Experiments and Results}

\subsection{Implement Details}

\subsubsection{Train-test Splitting} 

All experiments are conducted on StableDB. We follow the common practice of database splitting by leaving out 60\% for training, 20\% for validation, and 20\% for testing. As one split may cause bias when training a deep-learning-based model, we randomly split it ten times, and use the average results for performance comparison. 

\subsubsection{Training Protocol}

We utilize RAFT~\cite{teed2020raft} for optical flow estimation in preparation, as it is considered the SOTA model in optical estimation. For the flow feature branch in the feature extraction module, we utilize 3D ResNet-18 as the backbone for efficiency. For the blur feature branch, we use the encoder module and 6 Intra-SA and Inter-SA blocks in Stripformer~\cite{tsai2022stripformer}. The Intra-SA and Inter-SA blocks leverage horizontal and vertical strip-wise features to extract blurred patterns with different orientations and magnitudes based on attention mechanisms. Before training, we load the pre-trained checkpoint of Stripformer trained on the RealBlur database~\cite{rim2020real}. The weights of the blur feature extractor are fixed during training. For the semantic feature extractor, we initialize the weights by loading the checkpoint of Swin-T pre-trained on the ImageNet-1K database. Weights of Swin-T are later fine-tuned on StableDB. Other parts of StableVQA are randomly initialized. 

During training, we randomly sample a clip containing $N=32$ frames with the time interval $\tau=2$ from one video sequence. During validation and testing, we sample 4 clips under the same strategy and predict stability scores separately. The final score is obtained by averaging. All input video frames are resized into $224 \times 224$ for all training, validation, and testing. Following~\cite{wu2022fast}, we use Adam optimizer initialized by learning rate $0.0001$ for the Swin-T backbone and $0.001$ for the rest of the model. The learning rate decays under a cosine scheduler from 1 to 0. We train StableVQA for 30 epochs under a batch size of 4 on a server with one NVIDIA GeForce RTX 3090. 

\subsubsection{Loss Function}

We use differentiable Pearson Linear Correlation Coefficient (PLCC)~\cite{wu2022fast} and rank loss~\cite{sun2022deep} as loss function. PLCC is a common criterion used for evaluating the correlation between sequences, while the rank loss is introduced to help the model distinguish the relative quality of videos better. The differentiable PLCC loss is defined by:






\begin{equation}
    L = L_{plcc} + \lambda \cdot L_{rank},
\end{equation}
where $\lambda$ is a hyper-parameter for balancing, and is set to $0.3$ during training. 

\begin{figure*}[]
\centering
\subfigcapskip=9pt
\subfigbottomskip=-1pt
\subfigure[StableVQA]{
\begin{minipage}[t]{0.33\textwidth}
\centering
\begin{overpic}[width=0.9\textwidth]{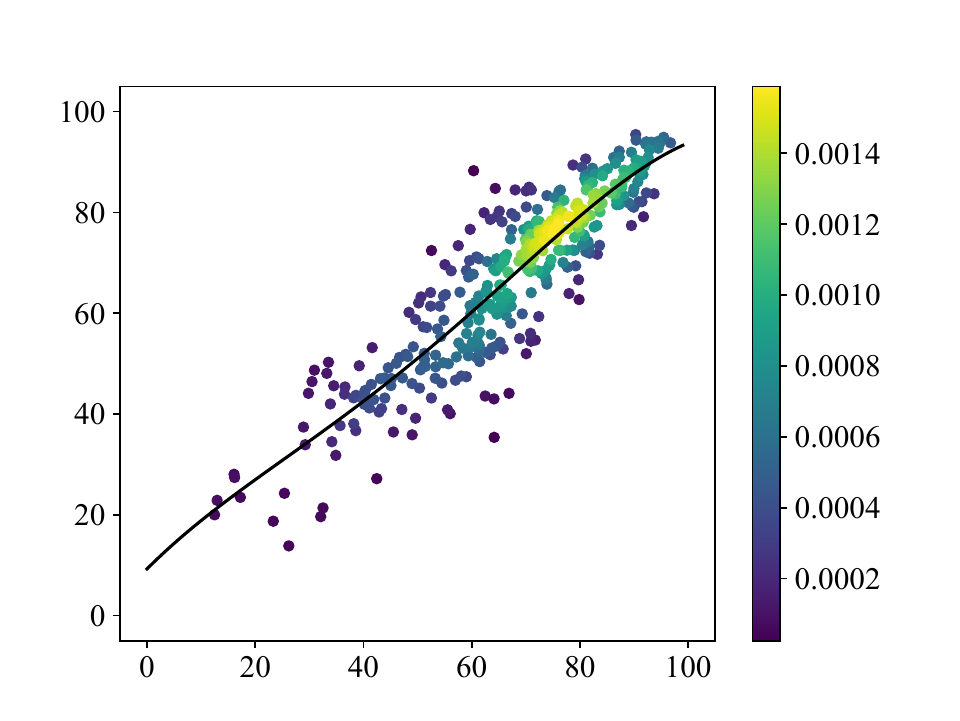}
\put(40, -5){\small{Predicted}}
\put(-2, 33){\rotatebox{90}{\small{MOS}}}
\end{overpic}
\end{minipage}%
}%
\subfigure[ITF]{
\begin{minipage}[t]{0.33\textwidth}
\centering
\begin{overpic}[width=0.9\textwidth]{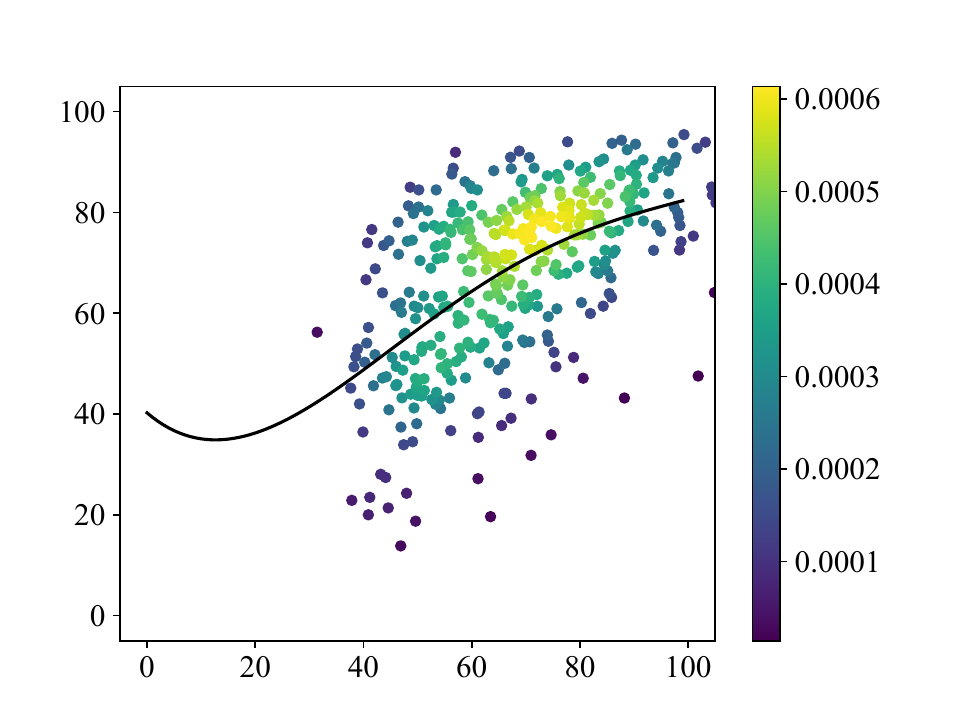}
\put(40, -5){\small{Predicted}}
\put(-2, 33){\rotatebox{90}{\small{MOS}}}
\end{overpic}
\end{minipage}%
}%
\subfigure[Stability Score]{
\begin{minipage}[t]{0.33\textwidth}
\centering
\begin{overpic}[width=0.9\textwidth]{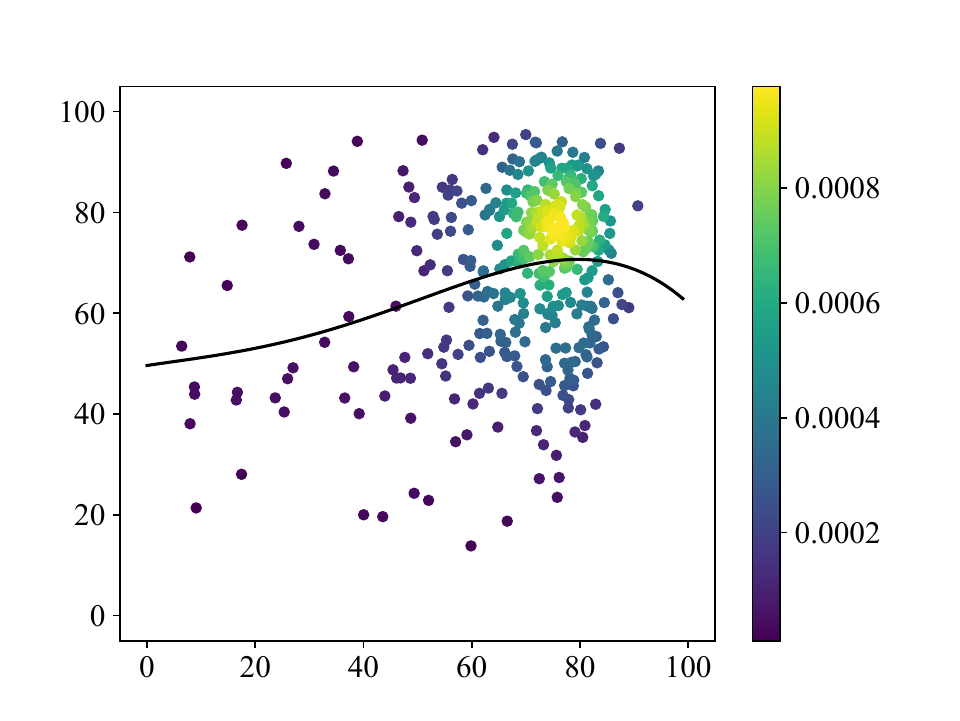}
\put(40, -5){\small{Predicted}}
\put(-2, 33){\rotatebox{90}{\small{MOS}}}
\end{overpic}
\end{minipage}
}\\%
\subfigure[VSFA]{
\begin{minipage}[t]{0.33\textwidth}
\centering
\begin{overpic}[width=0.9\textwidth]{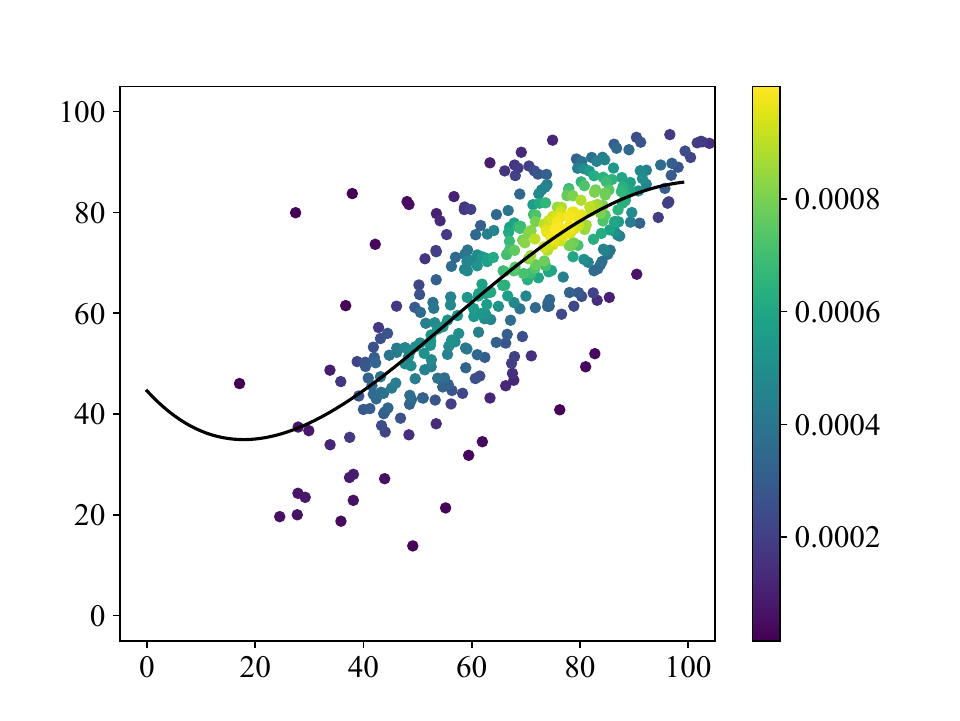}
\put(40, -5){\small{Predicted}}
\put(-2, 33){\rotatebox{90}{\small{MOS}}}
\end{overpic}
\end{minipage}
}%
\subfigure[SimpleVQA]{
\begin{minipage}[t]{0.33\textwidth}
\centering
\begin{overpic}[width=0.9\textwidth]{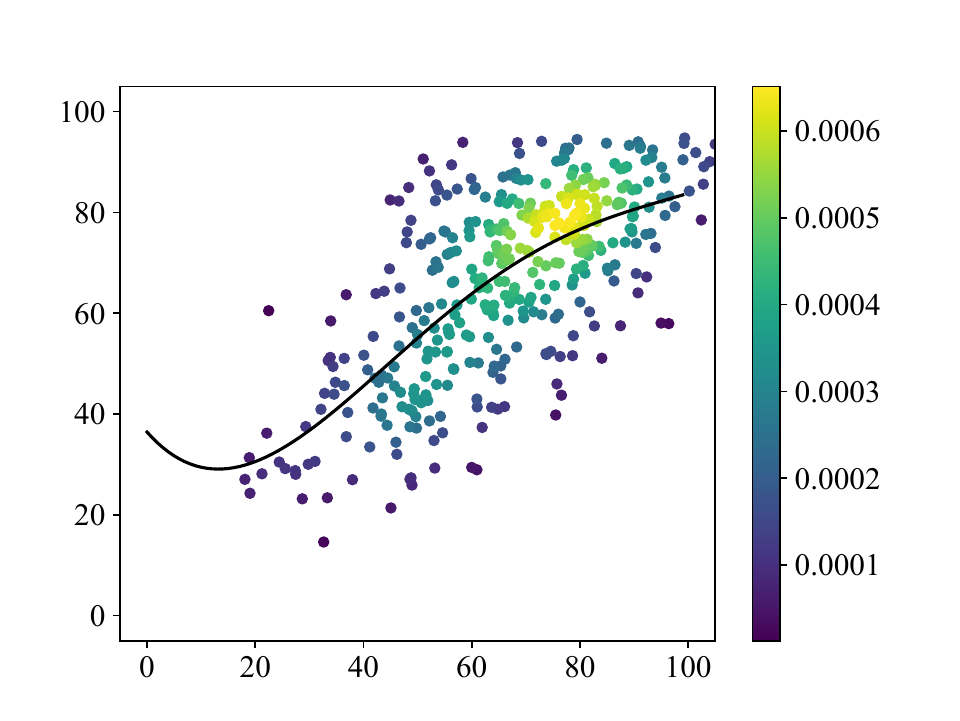}
\put(40, -5){\small{Predicted}}
\put(-2, 33){\rotatebox{90}{\small{MOS}}}
\end{overpic}
\end{minipage}
}%
\subfigure[FAST-VQA]{
\begin{minipage}[t]{0.33\textwidth}
\centering
\begin{overpic}[width=0.9\textwidth]{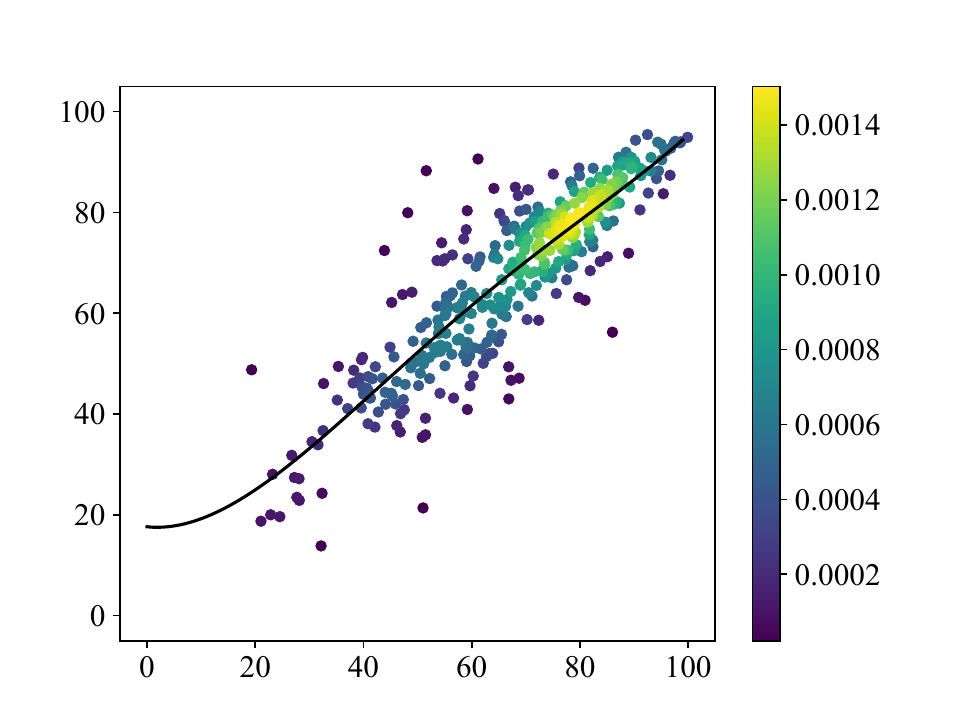}
\put(40, -5){\small{Predicted}}
\put(-2, 33){\rotatebox{90}{\small{MOS}}}
\end{overpic}
\end{minipage}
}%
\centering
\caption{Scatter plots of the predicted scores vs. MOSs. The curves are obtained by a four-order polynomial nonlinear fitting. Scatter points are distinguished by intensity using colors from dark to bright.}
\label{fig:performance}
\end{figure*}
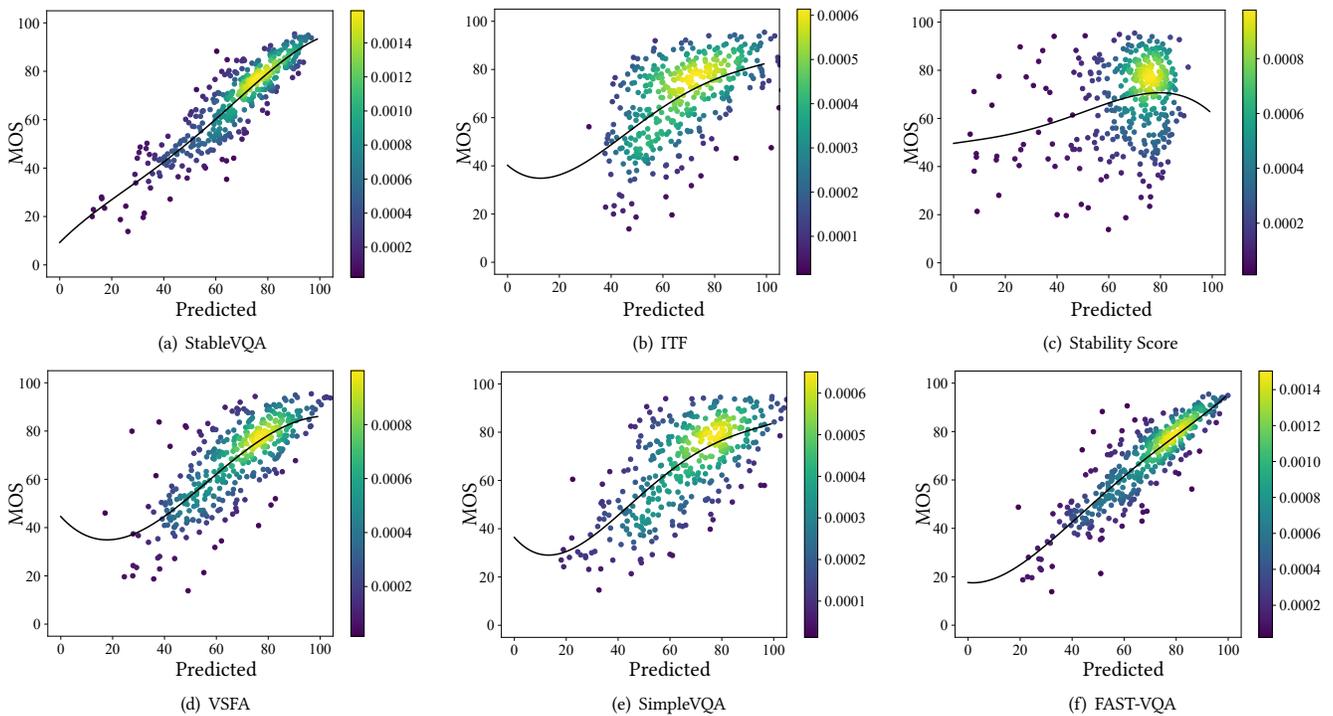

\subsubsection{Evaluation Metrics}

Besides PLCC, we include Spearman’s rank-order correlation coefficient (SROCC), Kendall’s rank-order correlation coefficient (KRCC), and root mean square error (RMSE) as performance criteria. 
Better models should have larger SROCC, KRCC, PLCC and smaller RMSE.  Before calculating the PLCC, we follow the same procedure in~\cite{video2000final} to map the objective score to the subject score using a four-parameter logistic function.

\begin{table}[]
\centering
\caption{Performance comparison on stabilization subset in LIVE-Qualcomm database.}
\renewcommand{\arraystretch}{1.2}
\begin{tabular}{c|cccc}
\toprule
\multirow{2}{*}{model} & \multicolumn{4}{c}{\begin{tabular}[c]{@{}c@{}}LIVE-Qualcomm~\cite{ghadiyaram2017capture}\\ (stabilization)\end{tabular}} \\ \cline{2-5} 
                        & SROCC $\uparrow$                                        & PLCC $\uparrow$  & KRCC $\uparrow$ & RMSE $\downarrow$                                     \\ \hline
ITF~\cite{battiato2007sift}                     &  -0.1969                                             & -0.2267                   & -0.1229            &  17.2062          \\
Stability Score~\cite{liu2013bundled}                     & 0.1789                                              & 0.1606                &  0.1266            & 14.233             \\
VSFA~\cite{li2019quality}                   &   0.3936
                                            &   0.3639
         &   0.3109
             &  26.7559
               \\
SimpleVQA~\cite{sun2022deep}              &  0.049
                                             &  0.164
               &   0.042
              &   10.7756
        \\
FAST-VQA~\cite{wu2022fast}               &    0.5661                                           &     0.5842           &   \textbf{0.4454}            &     9.9619      \\
\textbf{StableVQA}     &        \textbf{0.5815}                                       &      \textbf{0.6519}       &    0.4219     &   \textbf{9.1148}                      \\ \bottomrule 
\end{tabular}
\label{tab:live-qualcomm}
\end{table}

\subsection{Performance Comparison}

\subsubsection{Reference Algorithms}

We compare StableVQA with the following no-reference VQA-S and VQA algorithms:

\begin{itemize}
    \item VQA-S: ITF~\cite{battiato2007sift}, Stability Score~\cite{liu2013bundled}.
    \item VQA: VSFA~\cite{li2019quality}, SimpleVQA~\cite{sun2022deep}, BVQA~\cite{li2022blindly}, FAST-VQA~\cite{wu2022fast}, FAST-VQA + flow feature, FAST-VQA + blur feature.
\end{itemize}

For both VQA-S and VQA models, we calculate metrics between the predicted stability scores and MOSs on ten splits of StableDB and average to get final results. 

\subsubsection{Camprison with SOTA models}

Table~\ref{tab:performance} shows the performance comparison between StableVQA and reference algorithms, all trained on StableDB. Experimental results show that the proposed model has the best performance as it obtains higher correlation and lower error with subjective MOS. Figure~\ref{fig:performance} further shows the distribution of part models, where the data in the 10th test split is used. The horizontal axis in figure~\ref{fig:performance} represents the predicted scores, while the vertical axis represents the MOSs. 

ITF, VSFA, and SimpleVQA have similar performances on the proposed database, indicating they can generally reflect stability in videos. The Stability Score has the poorest performance. It is mainly caused by the following reasons: 1) Stability score uses the ratio between energy in low frequency against energy in total frequency. However, the demarcation of low frequency is manually set, which leads to uncertainty. 2) Trajectory estimation in stability score is based on feature point matching between adjacent frames, which always fails when facing severe shaking scenarios. 

FAST-VQA is considered the SOTA model in the VQA field. It gains effectiveness from the ``fragment'' sampling strategy and Swin Transformer backbone. Since video stabilization mainly focuses on temporal characteristics, ``fragment'' sampling's function becomes limited. Compared to FAST-VQA, the StableVQA adds the flow feature and blur feature branches to the Swin-T backbone, where the optical branch helps analyze camera movement in the temporal domain, while the blur branch detects blur effect in the spatial domain. These lead to an overall improvement in all four metrics. Besides, to investigate the effect of flow and blur features on existing models, we add the two types of features to FAST-VQA separately and test for their performance. Results show that they have slightly improve FAST-VQA's performance, but still fail to beat StableVQA.

\subsubsection{Qualitative Analysis}

\begin{figure}[]
\centering
\subfigcapskip=7pt
\subfigbottomskip=1pt
\subfigure[Origin: 27.9; Pr: 59.2; GlobalFlowNet: 60.3]{
\begin{minipage}[t]{\linewidth}
\centering
\begin{overpic}[width=\textwidth]{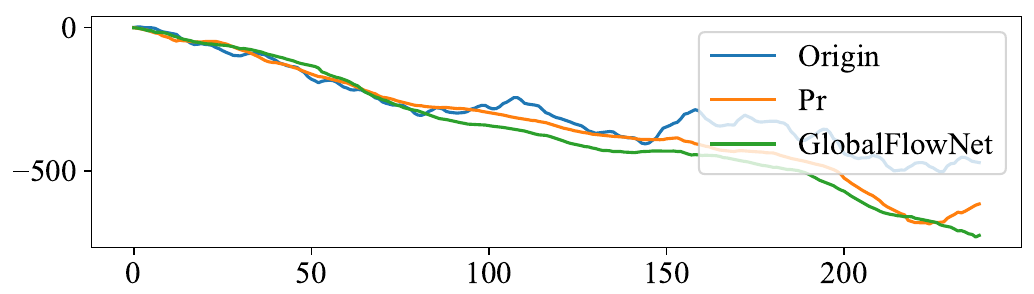}
\put(40, -3){\small{Frame Number}}
\put(-2, 15){\rotatebox{90}{\small{dy}}}
\end{overpic}
\end{minipage}%
}\\%
\subfigure[Origin: 20.0; Pr: 62.7; GlobalFlowNet: 57.3]{
\begin{minipage}[t]{\linewidth}
\centering
\begin{overpic}[width=\textwidth]{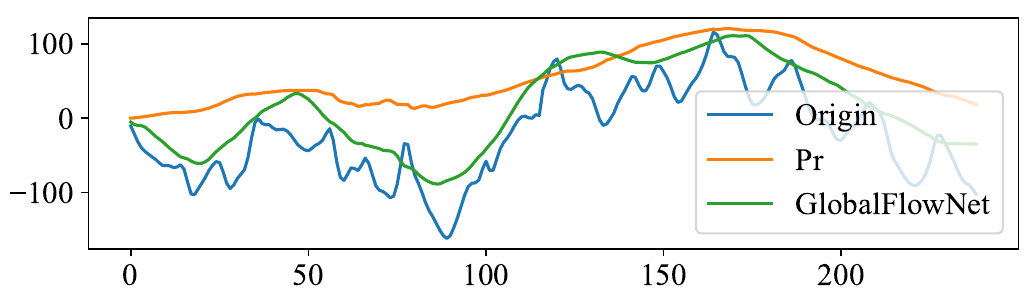}
\put(40, -3){\small{Frame Number}}
\put(-2, 15){\rotatebox{90}{\small{dy}}}
\end{overpic}
\end{minipage}
}\\%
\subfigure[Origin: 23.5; Pr: 66.1; GlobalFlowNet: 69.1]{
\begin{minipage}[t]{\linewidth}
\centering
\begin{overpic}[width=\textwidth]{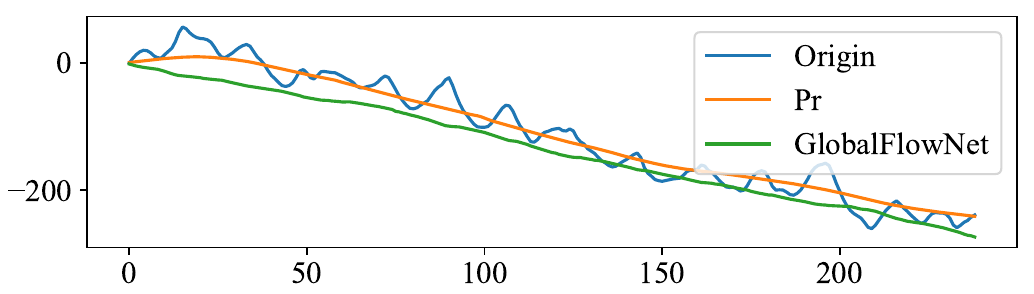}
\put(40, -3){\small{Frame Number}}
\put(-2, 15){\rotatebox{90}{\small{dy}}}
\end{overpic}
\end{minipage}%
}%

\centering
\caption{Trajectories of unstable and stabilized videos in vertical dimension. Predictions of StableVQA are shown in subtitles.  }
\label{fig:qualitative}
\end{figure}
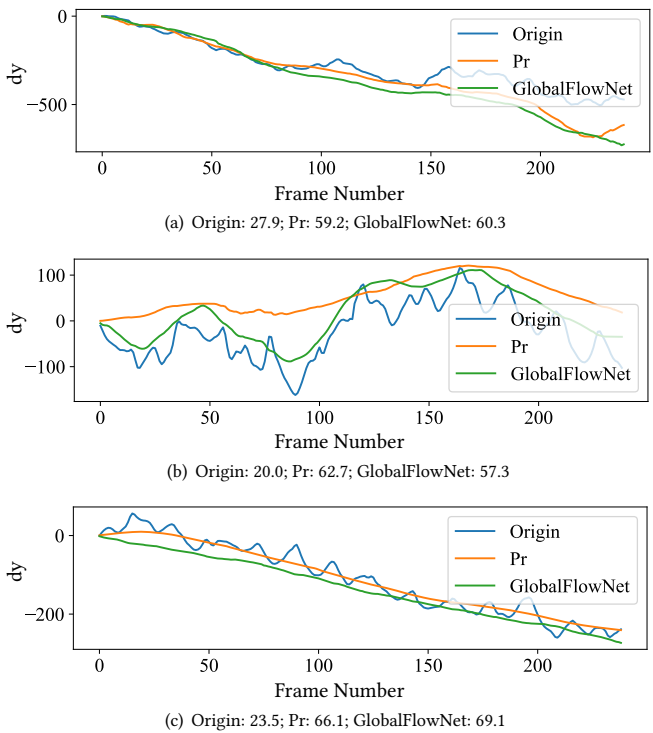

To further illustrate the effectiveness of StableVQA, we conduct a qualitative analysis by comparing predictions of the original unstable videos and the stable videos processed through stabilization algorithms. We use one software stabilization algorithm: the Warp Stabilizer in Adobe Premiere Pro 2022, and one deep-learning-based algorithm: GlobalFlowNet~\cite{james2023globalflownet} for video stabilization. In figure~\ref{fig:qualitative}, we demonstrate three examples with their trajectories in the vertical dimension. The blue curves show the unstable video trajectories, while the orange curves show the stabilized trajectories processed by Adobe Premiere Pro and the green for trajectories processed by GlobalFlowNet. We use StableVQA trained on the 10th split for stability prediction. The predicted scores for each unstable video are 27.9, 20.0, and 23.5 respectively. Corresponding Adobe-stabilized videos get 59.2, 62.7, and 66.1 respectively. GlobalFlowNet-stabilized videos get 60.3, 57.3, and 69.1 respectively. Qualitative analysis shows StableVQA is able to distinguish different degrees of video stability. Furthermore, the results give StableVQA the potential of evaluating the performance of video stabilization algorithms, improving its application value. 

\subsubsection{Cross-Database Validation}

To prove the generalization of StableVQA, a cross-database validation is needed. However, few databases have been proposed specifically focusing on the subjective experience of video stability. To the best of our knowledge, LIVE-Qualcomm~\cite{ghadiyaram2017capture} is the only one that has a subset of 35 videos with corresponding MOSs to measure video stability. We use the stabilization subset of the LIVE-Qualcomm database for cross-database validation. Since StableDB includes part of the videos in the LIVE-Qualcomm database, we remove those videos to avoid repeatability, leaving 26 videos for validation. Table~\ref{tab:live-qualcomm} further shows performance comparison over different models. All deep-learning-based models are trained on StableDB and tested on the LIVE-Qualcomm database. 

The testing results over LIVE-Qualcomm show consistency in the results on StableDB, as the proposed model achieves the best performance. Noticed that there is a distinct performance decrease on LIVE-Qualcomm compared with StableDB. The reason is that LIVE-Qualcomm includes several repeated scenes shot by different mobile devices. The stability within these videos is similar, but the MOSs can be various due to other types of distortions. Since the MOSs in StableDB are only considering video stability, the models trained on StableDB cannot distinguish differences in other distortion dimensions well.

\section{Ablation Study}
\label{ablation}

\begin{table}[]
\centering
\caption{The results of ablation studies. [Keys: of: flow feature; sf: semantic feature; bf: blur feature]}
\renewcommand{\arraystretch}{1.2}
\begin{tabular}{c|cc|cc}
\toprule
\multirow{2}{*}{model} & \multicolumn{2}{c|}{Validation}  & \multicolumn{2}{c}{Testing}\\ \cline{2-5} 
                        & SROCC $\uparrow$                                        & PLCC $\uparrow$ & SROCC $\uparrow$                                        & PLCC $\uparrow$                                       \\ \hline
of                     &    0.8899                                           &    0.8999       &      0.8857      &   0.8979                               \\
sf                      &    0.9067                                           &   0.9143             &    0.9041            &    0.9132         \\
bf                    &   0.3137                                            &   0.3125       &     0.2535           &   0.2386                \\
of+sf               &     0.9097                                          &     0.9145       &     0.9105           &    0.9176             \\
bf+sf                &    0.9047                                           &   0.9155        &    0.8978            &   0.9139              \\
of+bf                &    0.8756                                           &   0.8840        &    0.8810            &   0.8887              \\
\textbf{of+sf+bf}                    &    \textbf{0.9102}                                          &   \textbf{0.9161}            &     \textbf{0.9118}       &   \textbf{0.9187}           \\ \bottomrule 
\end{tabular}
\label{tab:ablation}
\end{table}

To investigate the effectiveness of flow, semantic, and blur features extracted in StableVQA, we thoroughly conduct the ablation studies of our StableVQA. First, we evaluate model performance using one type of feature separately. Afterward, we combine features in pairs. Finally, the StableVQA concatenates all features. All experiments are conducted on the StableDB. Table~\ref{tab:ablation} shows the detailed results. 

When using only one branch of features, the semantic feature performs the best, while the blur feature is the worst. The reason why video stability cannot be assessed with the blur feature is that video stability is decided by camera movement in the temporal domain. Since the blur feature only analyzes the blur effect in video frames in the spatial domain, it is difficult to analyze video characteristics temporally. The flow feature is obtained by analyzing optical flows in the temporal domain. As a result, it has relatively high scores. When using two branches of features, the flow and semantic features combination performs best. And adding blur and flow features to semantic feature has a similar performance. Finally, using all three branches of features achieves the best performance, indicating the effectiveness of all types of features.




\section{Conclusion}

In this paper, we focus on giving an accurate evaluation of the stability of in-the-wild videos. For that purpose, we build a deep-learning-based model for video stability assessment, named StableVQA. For training such a model, we further propose StableDB, a large-scale unstable video database, including 1952 in-the-wild videos with corresponding subjective MOSs on the degree of video stability. Experimental results show StableVQA can better predict video stability under subjective judgment by beating former VQA-S and generic VQA models. Qualitative experiments also show StableVQA can benefit the performance evaluation of video stabilization algorithms, which improves its application value.

\begin{acks}
This work was supported in part by the Shanghai Pujiang Program under Grant 22PJ1406800, in part by the National Natural Science Foundation of China under Grant 62225112 and Grant 61831015, and in part by the China Postdoctoral Science Foundation under Grant 2023TQ0212.
\end{acks}

\bibliographystyle{ACM-Reference-Format}
\balance
\bibliography{samples/acmmm}

\appendix

\section{Details in Database Establishment}
\label{establish}

The proposed StableDB includes 1,952 diversely-shaky videos sourcing from KoNViD-1k~\cite{hosu2017konstanz}, V3C1~\cite{rossetto2019v3c}, LIVE-Qualcomm~\cite{ghadiyaram2017capture}, YouTube UGC~\cite{wang2019youtube}, NUS~\cite{liu2013bundled}, DeepStab~\cite{wang2018deep}, Selfie~\cite{yu2018selfie}, and our own shooting. However, since the aforementioned public databases were designed for general Video Quality Assessment (VQA) tasks 
, the videos may not be suitable for the Video Quality Assessment for Stability (VQA-S) task. In KoNViD-1k, videos have diverse degrees of distortions such as blur, stall, high contrast, low light, etc. Severe distortions in these dimensions will affect user judgment when scoring the stability of videos in the later subjective study. In V3C1, videos originate from online stream platforms. These videos normally have several transitions, which may lead to an ambiguous definition of stability. Besides, videos in V3C1 have various ranges of resolutions, from 176p-4K. Since we later uniformly resize video resolution to 720p, those videos with rather small resolution or with an aspect ratio other than $16:9$ need to be removed. In LIVE-Qualcomm, videos are captured using different mobile devices synchronized by a rig consisting of four phone holders and a metal rod. Such a strategy results in repeated scenes and similar stability in videos. As a result, we retain one video for each scenario. Videos in YouTube UGC have similar characteristics to videos in V3C1 since they originate from online platforms as well. Additionally, part of the videos in YouTube UGC is synthetic, which mismatches our target to evaluate stability in in-the-wild videos. A manual selection is conducted to remove the aforementioned problematic videos. 


\begin{figure}[]
    \centering
    \includegraphics[width=\linewidth]{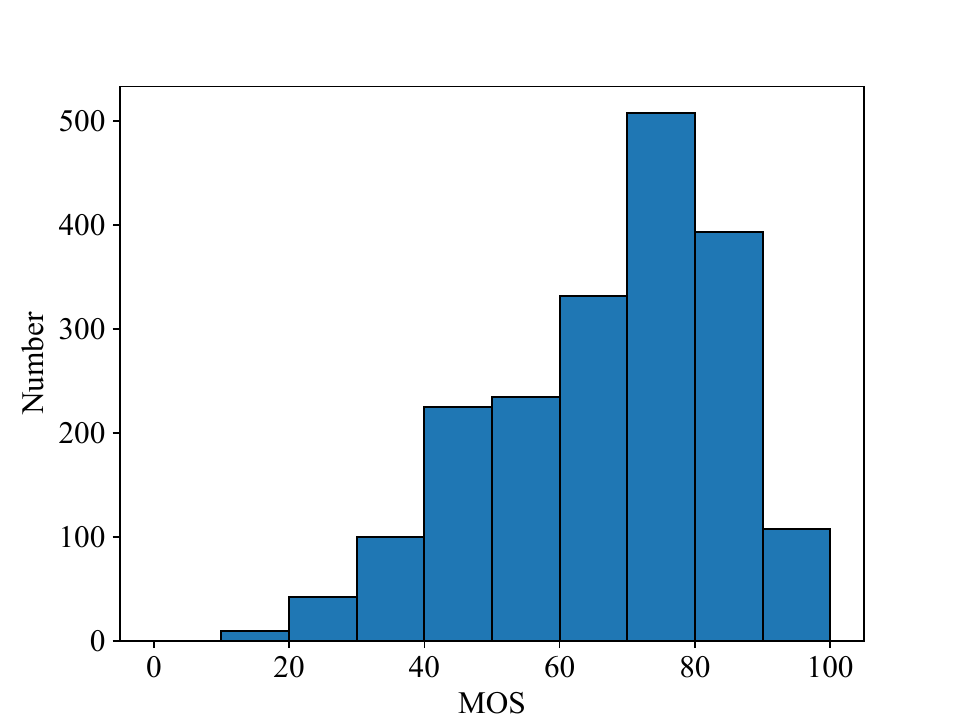}
    \caption{Distribution of MOSs on the stability of videos in StableDB.}
    \label{fig:mos_dis}
\end{figure}

\section{Details in Subjective Study}
\label{subjective}

Here we thoroughly introduce the workflow in the subjective study.

\textbf{Pilot Study:} As few studies have been conducted on subjective opinions of video shakiness, we conduct a pilot study to grasp a general understanding of stability distribution in StableDB. In the pilot study, 8 subjects are required to score 100 videos. 5 videos with the least range of ratings are selected as ``golden videos'' for quality control, as they are considered to have the highest consistency among all subjects. We also select 5 videos with a wide range of MOSs for training. 

\textbf{Introduction:} At the beginning, each subject will read a brief description of the overall settings and operations in the study. Subjects are told to focus on the shakiness of videos and ignore other types of distortions like blur, low light, high contrast, etc. Rating is required to be done after the video is fully played.


\textbf{Training:} In the training session, users are required to score on 5 videos with a diverse range of stability. Basically, there are five reference grades: bad, poor, fair, good, and excellent, representing from most unstable to most stable. The score ranges from 0 to 100 continuously, representing from bad to excellent. 

\textbf{Testing:} In the testing session, the whole database is randomly divided into two sessions. It takes roughly 3 hours to score each session. After the scoring, a data cleaning is conducted. For each video, we calculate the mean score and standard deviation. For each user, given a particular video, if the score from a user exceeds two standard deviations from the mean score of the video, we consider the score from this user as an outlier. If a user has outliers more than 5\% of the total number of videos, the user will be considered unreliable and his/her scores will be rejected. The remaining scores are averaged as the Mean Opinion Score (MOS) of each video. Figure~\ref{fig:mos_dis} shows the distribution of MOSs on the stability of videos in StableDB.

\section{Limitations and Future Work}
\label{analysis}

\begin{figure}[H]
    \centering
    \begin{overpic}[width=\linewidth]{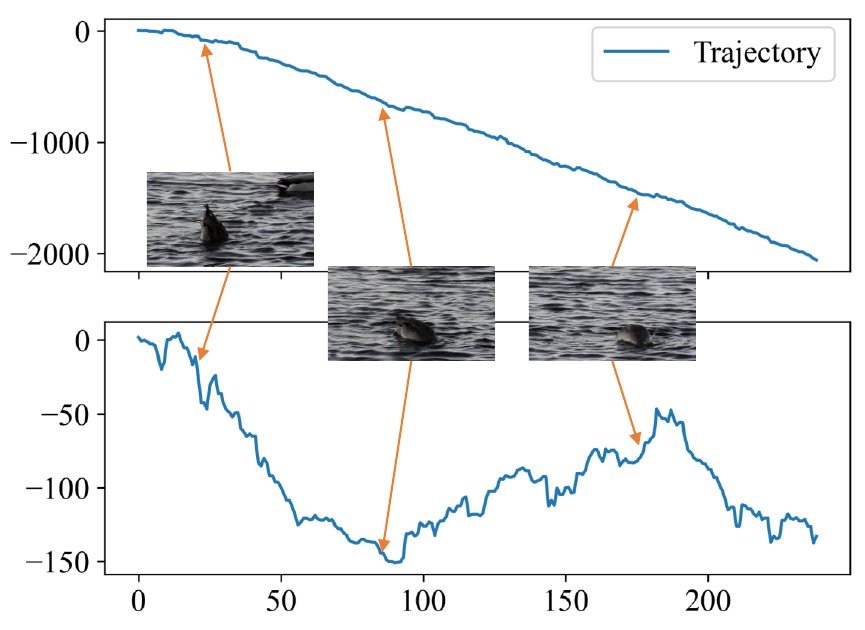}
    \put(40, -3){Frame Number}
    \put(-2, 20){\rotatebox{90}{\small{dy}}}
    \put(-2, 50){\rotatebox{90}{\small{dx}}}
    \end{overpic}
    \vspace{-2mm}
    \caption{Trajectory of a sample video on x and y axes, whose subjective MOS is 88.3 (range 0 to 100 from unstable to stable) while the prediction stability score from StableVQA is 61.9. }
    \label{fig:sample1}
\end{figure}

\begin{figure}[]
    \centering
    \begin{overpic}[width=\linewidth]{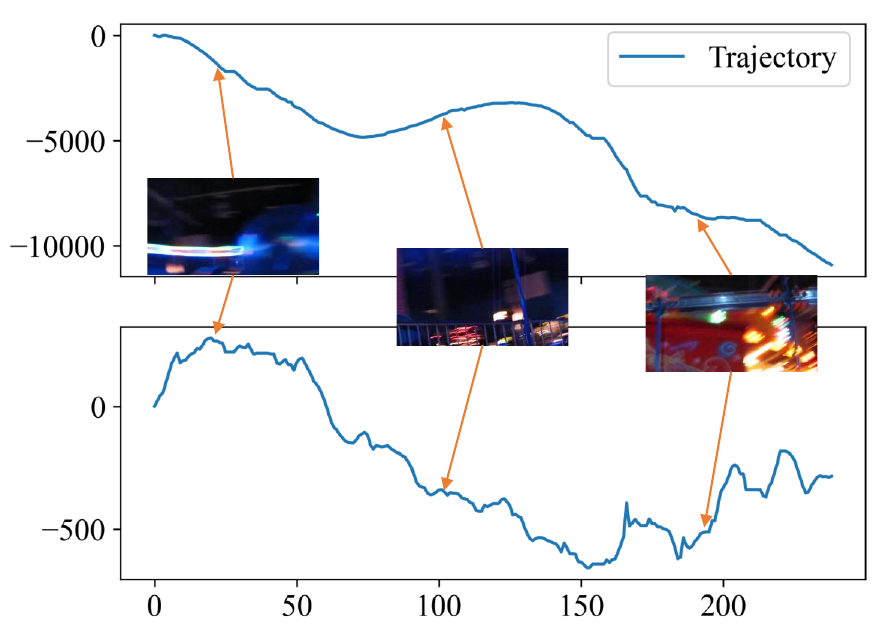}
    \put(40, -3){Frame Number}
    \put(-2, 20){\rotatebox{90}{\small{dy}}}
    \put(-2, 50){\rotatebox{90}{\small{dx}}}
    \end{overpic}
    \vspace{-2mm}
    \caption{Trajectory of another sample video on x and y axes, whose subjective MOS is 21.4 while the prediction stability score from StableVQA is 44.5.}
    \label{fig:sample2}
\end{figure}

In this section, we analyze the circumstances where the StableVQA fails to accurately predict the stability of videos. Figure~\ref{fig:sample1} and figure~\ref{fig:sample2} give two examples that are most representative in the 10th split test set. In figure~\ref{fig:sample1}, the video has a MOS of 88.3, indicating it has a relatively smooth camera movement. However, StableVQA gives a much lower prediction. In addition, trajectories in figure~\ref{fig:sample1} show that the video has a fast and uniform motion in the x direction, with severe shaking in the y direction. However, the real camera movement in the video is close to stationary. This deviation is because the scene in the video is about water waves and swimming ducks. Both water waves and ducks have relative motion with the camera, causing ambiguity for the model to make it believe the camera is moving rather than the objects in the scene. 

For the example video in figure~\ref{fig:sample2}, the MOS is 21.4, indicating the video has a rough camera trajectory. However, StableVQA predicts the stability score as 44.5. The camera in the video is on a rotating amusement facility, resulting in a quick rotation in the horizontal direction. Further, figure~\ref{fig:sample2} shows the trajectory in the x-axis is relatively smooth. However,  from the subjective perspective of the viewer, rotation under high velocity tends to amplify the jitter effect, making it has a relatively low MOS.

To conclude, StableVQA easily fails to predict accurately under the following circumstances: 1) scene in the video contains objects with similar characteristics performing relative motion that is inconsistent with the camera. 2) Mathematical trajectory smoothness is inconsistent with the subjective experience of stability, e.g., quick rotation. Both scenarios cause ambiguities in the model. FAST-VQA~\cite{wu2022fast} has a similar performance under these circumstances, where it predicts the stability of the video in figure~\ref{fig:sample1} with 55.5, and the video in figure~\ref{fig:sample2} with 49.7. Other models have even worse performances.

The future work lies in the reduction of computational complexity. A well-designed video stability model can be used for self-supervision in video stabilization algorithms. Besides, an interpretable algorithm is promising. With the proposed database, we hope to encourage the progress of research in video quality assessment and video stabilization fields.

\end{document}